\documentclass[10pt]{article} % For LaTeX2e
\usepackage[preprint]{tmlr}
\usepackage{amsmath,amsfonts,bm}

\def\eqref#1{equation~\ref{#1}}
\def\1{\bm{1}}

\DeclareMathAlphabet{\mathsfit}{\encodingdefault}{\sfdefault}{m}{sl}
\SetMathAlphabet{\mathsfit}{bold}{\encodingdefault}{\sfdefault}{bx}{n}

\usepackage{amsmath,amssymb}
\usepackage{amsthm}
\usepackage{hyperref}
\usepackage{url}
\usepackage{xurl}
\usepackage{tabularx}
\usepackage{float}
\usepackage{graphicx}
\usepackage{booktabs}
\usepackage{multirow}
\usepackage{xcolor}
\usepackage{algorithm}
\usepackage{algorithmic}
\usepackage{tikz}
\usetikzlibrary{arrows.meta,positioning,fit,backgrounds}
\usepackage{pgfplots}
\pgfplotsset{compat=1.17}
\usepgfplotslibrary{groupplots}

\newtheorem{theorem}{Theorem}
\newtheorem{proposition}{Proposition}

\newtheorem{corollary}{Corollary}
\theoremstyle{definition}
\newtheorem{definition}{Definition}
\theoremstyle{remark}
\newtheorem{remark}{Remark}

\newcommand{\proj}[1]{#1}

\newcommand{\Pref}{\tilde{P}_{\mathrm{ref}}}
\newcommand{\Aref}{A_{\mathrm{ref}}}
\newcommand{\sgraph}{s_{\mathrm{graph}}}
\newcommand{\ind}{\mathbb{1}}
\newcommand{\Ed}{\mathbb{E}}

\title{LogCDM: Interpretable Log Anomaly Detection via Behavioral Graph Drift and Adaptive Signal Fusion}

\author{
\name Suryansh Talukdar \email talusb01@pfw.edu \\
\addr Purdue University Fort Wayne
\AND
\name Taha Sarwat \email sarwt01@pfw.edu \\
\addr Purdue University Fort Wayne
\AND
\name Jin Soung Yoo \email jyoo@pfw.edu \\
\addr Purdue University Fort Wayne
}

\def\month{MM}  % Insert correct month for camera-ready version
\def\year{YYYY} % Insert correct year for camera-ready version
\def\openreview{\url{https://openreview.net/forum?id=XXXX}} % Insert correct link to OpenReview for camera-ready version

\begin{document}

\maketitle

% =====================================================================
\begin{abstract}
Log-based anomaly detection is increasingly dominated by large pretrained models, from BERT-scale encoders to 8B-parameter language models, which need GPUs at inference time and are hard to deploy on the monitoring nodes where logs are produced. We present LogCDM, a compact detector that scores each log session with three complementary signals: the next-event surprise of a small causal Transformer, the distance between a graph-attention embedding of the session's own transition graph and that of a normal reference graph, and a transition-drift statistic that we show is exactly a frequency-weighted $\chi^2$ divergence between transition kernels. The divergence view yields an asymptotic null distribution, a length-standardized score, and a per-edge decomposition that explains each alarm. A sigmoid gate fuses the three signals; under a shared-covariance Gaussian model this gate is the Bayes posterior, and its learned weights indicate which signal each dataset relies on. With \proj{8.1}--\proj{8.8}M parameters and no GPU at inference, LogCDM reaches F1 of \proj{0.984} on HDFS, \proj{0.870} on chronologically split BGL and \proj{0.850} on Thunderbird, within \proj{0.013}--\proj{0.12} of an 8.1B-parameter LLM detector evaluated under the same protocol and above all compact baselines on BGL. It scores a session in \proj{5.5}--\proj{10.5}\,ms on a CPU. Ablations show that the structural signals add \proj{0.025}--\proj{0.030} F1 on BGL and Thunderbird but almost nothing on HDFS, consistent with the learned fusion weights.
\end{abstract}

% =====================================================================
\section{Introduction}
\label{sec:intro}

Modern distributed systems, from Hadoop pipelines to supercomputers and cloud microservices, record their runtime behavior in system logs. Hardware faults, stalled jobs and cascading service failures all leave traces in the order and structure of logged events, and the volume of these logs rules out manual inspection \citep{he2021survey,landauer2023survey}. Log-based anomaly detection (LAD) asks whether a session or window of log events deviates from normal behavior, given historical logs and, in the semi-supervised setting, a small set of labeled anomalies.

Three properties make LAD hard. Raw log lines are free text whose surface form changes across software versions, so detectors first parse them into templates \citep{he2017drain}, and parsing errors bound what any downstream model can achieve \citep{shin2021theoretical}. Anomalies are rare, from about 3\% of HDFS sessions to about 1\% of Thunderbird windows, which makes threshold-based decisions fragile \citep{le2022log}. Finally, anomalies are heterogeneous: some insert a forbidden event, some change event counts, and some change how events follow one another across a whole session without introducing any unusual event locally \citep{meng2019loganomaly,li2024logs2graphs}.

\paragraph{Limitations of prior work.}
Most deep LAD methods score a session by how surprising its event sequence is under a model of normal sequences. DeepLog uses an LSTM next-event predictor \citep{du2017deeplog}; LogRobust and NeuralLog use semantic template embeddings with recurrent or Transformer encoders \citep{zhang2019logrobust,le2021neurallog}; LogBERT and LogFormer adapt masked-language-model pretraining \citep{guo2021logbert,guo2024logformer}. Recent detectors push accuracy further with large language models: LogLLM combines BERT with Llama-3-8B \citep{guan2024logllm}, CoLA routes hard cases from a small model to a fine-tuned 8B LLM \citep{zhu2025cola}, and LogLLaMA fine-tunes a LLaMA-2 architecture with reinforcement learning \citep{yang2025logllama}. This trend leaves two gaps.

First, \emph{cost}. LLM-based detectors need a GPU to score each session and gigabytes of memory to hold the model, while the nodes that most need monitoring (edge devices, cluster managers, on-premise servers) often have neither. Evaluation studies have long argued that time and resource cost belong next to accuracy when judging LAD systems \citep{le2022log,xiao2026protocol}, yet model size has grown by three orders of magnitude.

Second, \emph{structure and explanation}. Sequence surprise is computed locally, one next event at a time. It can miss sessions whose events are individually plausible but whose overall pattern of transitions has shifted, and it gives the operator a single number with no account of which transitions caused the alarm. Graph-based detectors model transition structure \citep{li2024logs2graphs}, but they usually replace sequence models instead of combining with them, and their scores lack a statistical interpretation.

\begin{figure}[t]
\centering
\begin{tikzpicture}[
    font=\small,
    box/.style={draw, rounded corners=2pt, align=center, minimum height=0.9cm, inner sep=3pt, fill=white},
    off/.style={box, fill=gray!10},
    sig/.style={box, fill=blue!6},
    arr/.style={-{Stealth[length=2mm]}, thick}
]
\node[box] (logs) {Raw logs};
\node[off, right=0.45cm of logs] (drain) {Drain3\\templates};
\node[off, right=0.45cm of drain] (emb) {MiniLM\\(once, offline)};
\node[off, below=0.55cm of drain] (ref) {Normal reference\\graph $\tilde A_{\mathrm{ref}}$};
\node[sig, right=1.1cm of emb] (tr) {Causal Transformer\\$p_e$};
\node[sig, below=0.25cm of tr] (gat) {Session GAT vs.\ ref.\\$s_{\mathrm{graph}}$};
\node[sig, below=0.25cm of gat] (stds) {STDS ($\chi^2$ drift)\\$z$};
\node[box, right=0.9cm of gat, fill=orange!10] (fuse) {Sigmoid gate\\$\hat y=\sigma(\mathbf{w}^{\top}\mathbf{s}+b)$};
\node[box, right=0.45cm of fuse] (out) {Alarm +\\top edges};
\draw[arr] (logs) -- (drain);
\draw[arr] (drain) -- (emb);
\draw[arr] (drain) -- (ref);
\draw[arr] (emb) -- (tr);
\draw[arr] (emb.south east) -- (gat.west);
\draw[arr] (ref.east) -- (gat.west);
\draw[arr] (ref.south east) -- (stds.west);
\draw[arr] (tr.east) -- (fuse.west);
\draw[arr] (gat) -- (fuse);
\draw[arr] (stds.east) -- (fuse.west);
\draw[arr] (fuse) -- (out);
\end{tikzpicture}
\caption{LogCDM overview. Grey boxes run once offline; blue boxes run per session on a CPU. The three scores are fused by a sigmoid gate, and the STDS term decomposes over transitions to explain each alarm.}
\label{fig:framework}
\end{figure}
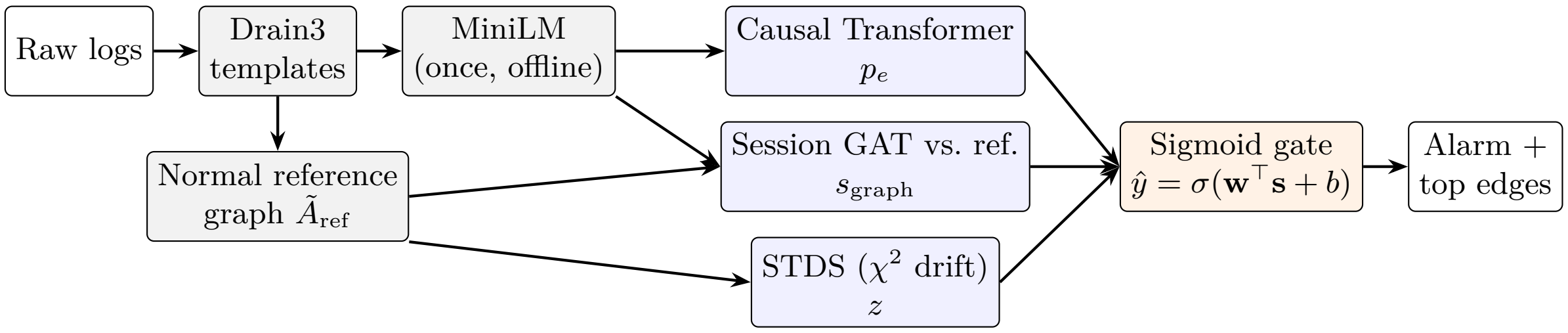

\paragraph{Our approach.}
LogCDM (Log Contextual Dual-pathway Model) combines three signals in one small model (Figure~\ref{fig:framework}). A 6-layer causal Transformer gives a sequence score. A graph attention network \citep{velickovic2018gat} embeds the session's own transition graph and compares it with an embedding of a normal reference graph. A closed-form Structural Transition Drift Score (STDS) compares the session's transition frequencies with the reference. A sigmoid gate fuses the three scores. Template embeddings are computed once offline, so the deployed model never calls a language model.

\paragraph{Contributions.}
\begin{enumerate}
    \item \textbf{A compact three-signal detector.} LogCDM has \proj{8.1}--\proj{8.8}M parameters, roughly 1{,}000$\times$ fewer than LLM-based detectors, and runs on a CPU (Section~\ref{sec:method}).
    \item \textbf{A statistical account of structural drift.} We prove that STDS equals a frequency-weighted $\chi^2$ divergence between transition kernels, and equivalently a weighted norm of the difference of random-walk Laplacians. This yields a total-variation lower bound, a bounded score under smoothing, an asymptotic $\chi^2$ null distribution, a length-standardized score, and a per-edge decomposition for explaining alarms (Section~\ref{sec:theory}).
    \item \textbf{An interpretable fusion gate.} Under a shared-covariance Gaussian model the gate is the Bayes posterior, so its standardized weights reflect each signal's class separation. We also show that a common way of feeding an adapted adjacency matrix to a GAT passes no gradient to the adapter, and give a support-preserving adapter that does (Section~\ref{sec:theory}).
    \item \textbf{A protocol-aware evaluation.} We evaluate on HDFS, BGL and Thunderbird with random and chronological splits, five seeds, AUC-PR, bootstrap intervals, CPU latency, nine ablations, label-noise tests and fusion-weight analysis (Section~\ref{sec:results}).
\end{enumerate}

% =====================================================================
\section{Related Work}
\label{sec:related}

\paragraph{Sequence models for log anomaly detection.}
DeepLog frames LAD as next-event prediction with an LSTM trained on normal sessions \citep{du2017deeplog}. LogAnomaly adds template semantics and count features \citep{meng2019loganomaly}, LogRobust uses FastText embeddings with an attention Bi-LSTM to tolerate evolving log statements \citep{zhang2019logrobust}, and \citet{lu2018cnn} use a CNN over embedded event sequences. NeuralLog skips parsing and encodes raw messages with BERT \citep{le2021neurallog}. LogBERT \citep{guo2021logbert} trains a small Transformer with masked log-key prediction and a hypersphere objective, and LogFormer \citep{guo2024logformer} pretrains on one system and adapts to another with adapters. LightLog targets edge devices with a temporal convolutional network \citep{wang2022lightlog}, and LogMamba uses state-space models with a frequency branch \citep{hu2026logmamba}. LogDA combines SBERT template vectors with dual attention for imbalanced data \citep{zhang2025logda}. These models score order and content but not session-level transition structure, and none gives a statistical meaning to its score.

\paragraph{Large language models for logs.}
LogLLM feeds BERT message embeddings into Llama-3-8B and reports the strongest results under a common protocol \citep{guan2024logllm}. CoLA lets a mixture-of-experts Bi-LSTM defer uncertain sessions to a fine-tuned Llama3-8B \citep{zhu2025cola}, and LogLLaMA fine-tunes a LLaMA-2 architecture with reinforcement learning \citep{yang2025logllama}. These methods set the accuracy frontier but require GPUs and billions of parameters at inference. LogCDM targets the opposite end of this trade-off.

\paragraph{Graph-based detection.}
Logs2Graphs converts sessions to attributed directed graphs and detects anomalies with a one-class digraph convolutional network, showing that transition topology carries signal beyond event order \citep{li2024logs2graphs}. We keep this insight but embed each session's graph next to a sequence model, compare it with a reference graph, and add a closed-form divergence with known statistical properties. Pearson-type tests on Markov transition counts go back to \citet{anderson1957markov}; our contribution is to connect that classical test to a learned detector, use it for length standardization and explanation, and fuse it with neural scores.

\paragraph{Parameter-efficient adaptation and score fusion.}
LoRA adapts frozen weights with low-rank updates \citep{hu2022lora}, and GraphLoRA brings low-rank adaptation to graph neural networks \citep{yang2024graphlora}. We use a low-rank, support-preserving correction of the reference adjacency matrix instead of the network weights. Our two-phase training, normal-only pretraining followed by a small labeled set, follows semi-supervised anomaly detection \citep{ruff2020deepsad,ruff2021unifying}, and our fusion gate is the discriminative counterpart of linear discriminant analysis \citep{hastie2009esl,ng2002discriminative}.

\paragraph{Lightweight and explainable detection.}
Compact detectors for constrained hardware include LightLog's temporal convolutions \citep{wang2022lightlog} and LogFormer's adapter tuning, which trains about 0.4M--1M parameters \citep{guo2024logformer}. Neither reports CPU latency together with accuracy under a chronological protocol, which is the comparison \citet{xiao2026protocol} argue for. Explanations for log anomalies are usually post hoc, for example subgraph explanations of a graph detector's output \citep{li2024logs2graphs}. In LogCDM the explanation is part of the score: the STDS term decomposes exactly over transitions, so the evidence behind an alarm is available without an extra explainer.

\paragraph{Evaluation practice.}
Random splits leak future information and inflate scores, while chronological splits lower them and can reorder methods \citep{le2022log,xiao2026protocol}. Common benchmarks also contain sessions that are trivially separable \citep{landauer2024critical}. We therefore report chronological splits, AUC-PR \citep{saito2015precision}, seed variance and bootstrap intervals, and separate same-protocol from different-protocol comparisons.

% =====================================================================
\section{Methodology}
\label{sec:method}

\subsection{Problem setup}

\begin{definition}[Sessions and vocabulary]
A parser $\pi$ maps each log line to a template in a finite vocabulary $\mathcal{V}=\{1,\dots,V\}$, with $0$ reserved for padding. A session is a sequence $x^{(i)}=(e_1^{(i)},\dots,e_T^{(i)})\in(\mathcal{V}\cup\{0\})^T$ grouped by block ID (HDFS) or by a fixed window of log lines (BGL, Thunderbird), with label $y^{(i)}\in\{0,1\}$.
\end{definition}

\begin{definition}[Leakage-free partition]
Sessions are split into disjoint sets $\mathcal{I}_{\mathrm{tr}},\mathcal{I}_{\mathrm{val}},\mathcal{I}_{\mathrm{te}}$, either chronologically or by stratified random sampling. The parser, vocabulary, embedding projection, reference graph, score normalization constants and decision threshold are all estimated from $\mathcal{I}_{\mathrm{tr}}$ and $\mathcal{I}_{\mathrm{val}}$ only. Templates first seen in $\mathcal{I}_{\mathrm{te}}$ map to an unknown token.
\end{definition}

The task is to learn a score $f:x\mapsto[0,1]$ and threshold $\tau^\star$ such that $\hat y=\ind[f(x)>\tau^\star]$ maximizes F1 on $\mathcal{I}_{\mathrm{val}}$; we report performance on $\mathcal{I}_{\mathrm{te}}$. Training uses normal sessions $\mathcal{S}_N\subset\mathcal{I}_{\mathrm{tr}}$ and a labeled fusion set $\mathcal{S}_F\subset\mathcal{I}_{\mathrm{tr}}$ containing a fraction of the training sessions with their labels.

\subsection{Preprocessing}

\paragraph{Parsing and embeddings.}
Drain3 \citep{he2017drain} is fitted on training logs. Each template's text is encoded once with all-MiniLM-L6-v2 \citep{reimers2019sbert,wang2020minilm} into $\mathbb{R}^{384}$ and mapped to $d=256$ dimensions, by PCA when the vocabulary is large enough (BGL, Thunderbird) and by a learned linear projection otherwise (HDFS has only \proj{39} templates). The result $E\in\mathbb{R}^{V\times d}$ initializes the Transformer's embedding table and the GAT node features. The sentence encoder is never called during training or inference.

\paragraph{Transition counts.}
For a session $i$ and maximum lag $L$, let
\begin{equation}
N_i(u,v)=\sum_{\ell=1}^{L}\sum_{t=1}^{T-\ell}\ind[e_t^{(i)}=u]\,\ind[e_{t+\ell}^{(i)}=v],\qquad n_i(u)=\sum_v N_i(u,v),\qquad n_i=\sum_u n_i(u).
\label{eq:counts}
\end{equation}
Let $U_i=\{u:n_i(u)>0\}$ be the source events the session visits, $\pi_i(u)=n_i(u)/n_i$ their empirical frequencies, and $P_i(v\mid u)=N_i(u,v)/n_i(u)$ the session's transition kernel. The reference counts $\Aref$ sum Eq.~\eqref{eq:counts} over $\mathcal{S}_N$, and $P_{\mathrm{ref}}$ is their row normalization.

\paragraph{Smoothed, coarsened reference.}
For each source $u$, let $\Omega_u=\operatorname{supp}P_{\mathrm{ref}}(\cdot\mid u)\cup\{\varnothing\}$, where $\varnothing$ pools all transitions never seen from $u$ in normal data. We define
\begin{equation}
\Pref(\omega\mid u)=(1-\lambda)\,P_{\mathrm{ref}}(\omega\mid u)+\lambda\,\ind[\omega=\varnothing],\qquad \omega\in\Omega_u,
\label{eq:smooth}
\end{equation}
with $\lambda=10^{-3}$, and map $P_i(\cdot\mid u)$ onto $\Omega_u$ in the same way. Every cell of $\Pref$ is then positive, and a novel transition counts as mass on $\varnothing$ instead of an infinite deviation.

\subsection{Three anomaly signals}

\paragraph{Sequence score.}
A causal Transformer encoder \citep{vaswani2017attention} with $N_{\mathrm{layer}}=6$ pre-norm layers, model width $d=256$, 8 heads and feed-forward width $d_{\mathrm{ffn}}=2048$ maps the embedded, positionally encoded session to contexts $C^{(i)}\in\mathbb{R}^{T\times d}$. A linear head predicts the next template, and the sequence score is the mean next-event cross-entropy over non-padding targets:
\begin{equation}
p_e^{(i)}=\frac{\sum_{t=1}^{T-1}\ind[e_{t+1}\neq0]\;\ell_{\mathrm{CE}}\big(W_{\mathrm{head}}C_t^{(i)},\,e_{t+1}\big)}{\sum_{t=1}^{T-1}\ind[e_{t+1}\neq0]}.
\label{eq:pe}
\end{equation}

\paragraph{Structural Transition Drift Score.}
STDS measures how far the session's transition kernel lies from the reference, weighted by how often each source event occurs:
\begin{equation}
s_i^2=\sum_{u\in U_i}\pi_i(u)\sum_{\omega\in\Omega_u}\frac{\big(P_i(\omega\mid u)-\Pref(\omega\mid u)\big)^2}{\Pref(\omega\mid u)}.
\label{eq:stds}
\end{equation}
Section~\ref{sec:theory} shows that $n_is_i^2$ is Pearson's statistic for the hypothesis that the session follows the reference kernel. Because its null mean grows with the number of cells and shrinks with session length, raw $s_i^2$ is biased toward short sessions. We therefore feed the fusion gate the standardized score
\begin{equation}
z_i=\frac{n_is_i^2-d_i}{\sqrt{2d_i}},\qquad d_i=\sum_{u\in U_i}\big(|\Omega_u|-1\big),
\label{eq:zstds}
\end{equation}
which has approximately zero mean and unit variance on normal sessions of any length (Corollary~\ref{cor:consistency}).

\paragraph{Session-graph score.}
Let $A_i$ be the session's transition-count matrix restricted to its visited events. A two-layer graph attention network with edge-weighted attention,
\begin{equation}
\alpha_{uv}=\frac{A[u,v]\exp\!\big(\mathrm{LeakyReLU}(\mathbf{a}^{\top}[Wh_u\,\Vert\,Wh_v])\big)}{\sum_{k}A[u,k]\exp\!\big(\mathrm{LeakyReLU}(\mathbf{a}^{\top}[Wh_u\,\Vert\,Wh_k])\big)},\qquad h_u'=\mathrm{ELU}\Big(\sum_v\alpha_{uv}Wh_v\Big),
\label{eq:gat}
\end{equation}
encodes $A_i$ into a mean-pooled, $\ell_2$-normalized embedding $g_i$. The same network encodes the adapted reference graph $\tilde A_{\mathrm{ref}}$ into $g_{\mathrm{ref}}$, and
\begin{equation}
\sgraph^{(i)}=\lVert g_i-g_{\mathrm{ref}}\rVert_2^2.
\label{eq:sgraph}
\end{equation}
Weighting attention by $A[u,v]$ (rather than using $A$ only to select neighbors) lets gradients reach the reference graph (Proposition~\ref{prop:adapter}).

\paragraph{Low-rank reference adapter.}
The reference graph is corrected multiplicatively on its existing edges:
\begin{equation}
\tilde A_{\mathrm{ref}}=\Aref\odot\big(\mathbf{1}\mathbf{1}^{\top}+\beta\tanh(PQ^{\top})\big),\qquad P,Q\in\mathbb{R}^{V\times r},\ r=4,\ \beta=0.1,
\label{eq:adapter}
\end{equation}
with $Q$ initialized to zero. The adapter has $2Vr$ parameters (\proj{312} on HDFS, \proj{10{,}400} on BGL), leaves the reference unchanged at initialization, keeps its support, and keeps every weight within $[1-\beta,1+\beta]$ of its original value.

\subsection{Fusion and training}

Each score is min-max normalized with training-split constants, giving $\mathbf{s}^{(i)}=(\tilde p_e^{(i)},\tilde s_{\mathrm{graph}}^{(i)},\tilde z_i)^{\top}$, and
\begin{equation}
\hat y^{(i)}=\sigma\big(\mathbf{w}^{\top}\mathbf{s}^{(i)}+b\big),\qquad \mathbf{w}=(w_a,w_b,w_c)^{\top}.
\label{eq:fusion}
\end{equation}

\paragraph{Phase 1: normal-only pretraining.}
Only the Transformer is trained, on $\mathcal{S}_N\setminus\mathcal{S}_F$, to minimize the mean of $p_e$ (Eq.~\ref{eq:pe}). The checkpoint with the lowest cross-entropy on held-out normal validation sessions is kept.

\paragraph{Phase 2: labeled fusion.}
All parameters are trained on $\mathcal{S}_F$ with a class-balanced sampler and
\begin{equation}
\mathcal{L}=\mathcal{L}_{\mathrm{BCE}}+\lambda_1\mathcal{L}_{\mathrm{rec}}+\lambda_2\mathcal{L}_{\mathrm{cont}},\qquad \lambda_1=0.15,\ \lambda_2=0.10,
\label{eq:loss}
\end{equation}
where $\mathcal{L}_{\mathrm{BCE}}$ is binary cross-entropy on $\hat y$ with label smoothing $0.1$, $\mathcal{L}_{\mathrm{rec}}=\Ed_{y=0}[p_e]$ keeps the Phase 1 sequence model close to its normal prior, and $\mathcal{L}_{\mathrm{cont}}=\max\big(0,\gamma-(\Ed_{y=1}[\sgraph]-\Ed_{y=0}[\sgraph])\big)$ with $\gamma=0.5$ is a margin term in the spirit of \citet{hadsell2006contrastive}. After each epoch the threshold $\tau^\star$ maximizing validation F1 is computed over quantiles of validation scores, and the checkpoint with the best validation F1 is kept. Algorithm~\ref{alg:train} summarizes training.

\begin{algorithm}[t]
\caption{LogCDM training}
\label{alg:train}
\let\AND\relax % tmlr.sty and algorithmic both define \AND; reset locally inside this float
\begin{algorithmic}[1]
\REQUIRE Training logs; labeled fusion set $\mathcal{S}_F$; validation set; lag $L$; smoothing $\lambda$
\STATE Fit Drain3 on training logs; build vocabulary; compute $E$ once with MiniLM
\STATE Build $\Aref$ from normal training sessions; form $\Pref$ by Eq.~\eqref{eq:smooth}
\STATE \textbf{Phase 1:} train the Transformer on $\mathcal{S}_N\setminus\mathcal{S}_F$ to minimize $\Ed[p_e]$; keep best normal-validation checkpoint
\STATE Fit min-max constants for $p_e,\sgraph,z$ on training sessions
\STATE \textbf{Phase 2:} initialize $Q\leftarrow0$; train all parameters on $\mathcal{S}_F$ with Eq.~\eqref{eq:loss}
\STATE \quad after each epoch: refit normalization constants; choose $\tau^\star$ on validation; keep best checkpoint
\STATE Cache $g_{\mathrm{ref}}$; \textbf{return} model, $\tau^\star$, constants
\end{algorithmic}
\end{algorithm}

\subsection{Inference}
At deployment, $g_{\mathrm{ref}}$ and $\Pref$ are cached. Each session needs one Transformer pass, a GAT pass over at most $T$ nodes and the closed-form STDS. The cost is dominated by the Transformer, $O\big(N_{\mathrm{layer}}(Td\,d_{\mathrm{ffn}}+T^2d)\big)$ per session, and the STDS costs $O(n_i)$ with sparse counts.

% =====================================================================
\section{Theoretical Analysis}
\label{sec:theory}

This section states the properties of the structural score, the fusion gate and the adapter. Full proofs are in Appendix~\ref{app:proofs}.

\subsection{STDS as a divergence between transition kernels}

For distributions $p,q$ on a finite set with $q>0$, write $\chi^2(p\Vert q)=\sum_\omega(p(\omega)-q(\omega))^2/q(\omega)$ and $\mathrm{TV}(p,q)=\tfrac12\sum_\omega|p(\omega)-q(\omega)|$. The random-walk Laplacian of a row-stochastic matrix $P$ is $\mathcal{L}=I-P$.

\begin{theorem}[STDS is a weighted $\chi^2$ divergence]
\label{thm:stds}
Let $\tilde p_{\min}=\min_{u,\omega}\Pref(\omega\mid u)>0$. For every session $i$:
\begin{enumerate}
    \item[(a)] \emph{Divergence form.} $s_i^2=\sum_{u\in U_i}\pi_i(u)\,\chi^2\big(P_i(\cdot\mid u)\,\Vert\,\Pref(\cdot\mid u)\big)$. Hence $s_i\ge0$, with equality if and only if $P_i(\cdot\mid u)=\Pref(\cdot\mid u)$ for every visited $u\in U_i$.
    \item[(b)] \emph{Laplacian form.} With $\mathcal{L}_i=I-P_i$ and $\tilde{\mathcal{L}}_{\mathrm{ref}}=I-\Pref$ on the coarsened outcome spaces, $s_i^2=\big\Vert\Pi_i^{1/2}(\mathcal{L}_i-\tilde{\mathcal{L}}_{\mathrm{ref}})\oslash\Pref^{1/2}\big\Vert_F^2$, where $\Pi_i=\mathrm{diag}(\pi_i)$ and $\oslash$, $\cdot^{1/2}$ act entrywise.
    \item[(c)] \emph{Total-variation bound.} $4\sum_{u\in U_i}\pi_i(u)\,\mathrm{TV}^2\big(P_i(\cdot\mid u),\Pref(\cdot\mid u)\big)\le s_i^2$.
    \item[(d)] \emph{Boundedness.} $s_i^2\le1/\tilde p_{\min}-1$, so no single novel transition can make the score unbounded.
    \item[(e)] \emph{Edge decomposition.} $s_i^2=\sum_{(u,\omega)}c_i(u,\omega)$ with $c_i(u,\omega)=\pi_i(u)\,(P_i(\omega\mid u)-\Pref(\omega\mid u))^2/\Pref(\omega\mid u)\ge0$.
\end{enumerate}
\end{theorem}

\textit{Proof sketch.} Part (a) regroups Eq.~\eqref{eq:stds} by source. Part (b) holds because $\mathcal{L}_i-\tilde{\mathcal{L}}_{\mathrm{ref}}=\Pref-P_i$ entrywise, including the diagonal, so the Laplacian difference and the kernel difference have the same squared entries; this is why the earlier unnormalized-Laplacian formulation failed on self-loops and the random-walk form does not. Part (c) applies Cauchy--Schwarz, $\sum_\omega|p-q|\le\sqrt{\chi^2(p\Vert q)}$, row by row. Part (d) uses $\chi^2(p\Vert q)=\sum_\omega p^2/q-1\le1/\min q-1$ and $\sum_u\pi_i(u)=1$. Part (e) is the definition. \hfill$\square$

Part (e) is what makes LogCDM's alarms interpretable: the contributions $c_i(u,\omega)$ rank the transitions that drove a session's score (Section~\ref{sec:interp}).

The next result formalizes the blind spot described in Section~\ref{sec:intro}. Many sequence detectors, DeepLog with a top-$g$ rule being the clearest case \citep{du2017deeplog}, flag a session only if some observed transition is unlikely under the normal model. STDS also reacts when every transition is individually normal but their frequencies have shifted.

\begin{proposition}[Frequency shifts invisible to support-based rules]
\label{prop:blind}
Call a detector support-based if it flags a session only when the session contains a transition $(u,v)$ with $P_{\mathrm{ref}}(v\mid u)=0$; a top-$g$ next-event rule with $g$ at least the maximum out-degree of the reference graph is an example. Let $u$ have at least two reference successors and let $v^\star$ be its least likely one, with $q^\star=\Pref(v^\star\mid u)$. Any session that visits $u$ with frequency $\pi_i(u)$, always moves from $u$ to $v^\star$, and otherwise follows the reference is never flagged by a support-based rule, yet
\[
s_i^2\ \ge\ \pi_i(u)\Big(\frac{1}{q^\star}-1\Big).
\]
\end{proposition}

\textit{Proof sketch.} Every transition lies in the reference support, so a support-based rule stays silent. Row $u$ of the session is the point mass $\delta_{v^\star}$, and $\chi^2(\delta_{v^\star}\Vert q)=(1-q^\star)^2/q^\star+\sum_{\omega\neq v^\star}q(\omega)=1/q^\star-1$; the other rows contribute nonnegative terms. \hfill$\square$

For a successor seen in 1\% of normal transitions from $u$, the bound is $99\,\pi_i(u)$, close to the maximum allowed by Theorem~\ref{thm:stds}(d). Repetition bursts of an otherwise normal event, as in the example of Section~\ref{sec:interp}, have this form. A learned sequence model can still assign them some surprise, but nothing in its objective guarantees it; STDS reacts by construction.

\subsection{Null distribution and length standardization}

\begin{proposition}[Asymptotic null distribution]
\label{prop:null}
Let $L=1$. Suppose a session is generated by a Markov chain whose transition kernel, after coarsening to $\Omega_u$, equals $\Pref$, and let $\min_{u\in U_i}n_i(u)\to\infty$ with $U_i$ fixed. Then
\[
n_is_i^2=\sum_{u\in U_i}\sum_{\omega\in\Omega_u}\frac{\big(N_i(u,\omega)-n_i(u)\Pref(\omega\mid u)\big)^2}{n_i(u)\Pref(\omega\mid u)}\ \xrightarrow{d}\ \chi^2_{d_i},\qquad d_i=\sum_{u\in U_i}\big(|\Omega_u|-1\big).
\]
\end{proposition}

The identity follows from $\pi_i(u)=n_i(u)/n_i$; the limit is the Pearson goodness-of-fit theorem for Markov chains of \citet{anderson1957markov}.

\begin{corollary}[Consistency and length standardization]
\label{cor:consistency}
Under the conditions of Proposition~\ref{prop:null}, $\Ed[s_i^2]\approx d_i/n_i$, so raw STDS is larger for short sessions even when they are normal, whereas $z_i$ in Eq.~\eqref{eq:zstds} has asymptotic mean $0$ and variance $1$ for any length. If instead the session follows an ergodic chain with kernel $P\neq\Pref$ on some visited row, so that $\delta=\sum_u\bar\pi(u)\chi^2(P(\cdot\mid u)\Vert\Pref(\cdot\mid u))>0$ for the stationary frequencies $\bar\pi$, then $s_i^2\to\delta$ almost surely, $z_i\to\infty$, and the test $\ind[n_is_i^2>\chi^2_{d_i,1-\alpha}]$ has power tending to one.
\end{corollary}

\begin{remark}
For lags $L>1$ the counts overlap and are no longer conditionally multinomial, and for short sessions with large vocabularies the $\chi^2$ approximation is loose. In these regimes we use $z_i$ only as a length-normalized score and let the fusion gate calibrate it; we do not report $p$-values.
\end{remark}

\subsection{Fusion as a Bayes classifier}

\begin{proposition}[Fusion gate under a Gaussian score model]
\label{prop:fusion}
Suppose $\mathbf{s}\mid y\sim\mathcal{N}(\boldsymbol{\mu}_y,\Sigma)$ for $y\in\{0,1\}$ with shared $\Sigma\succ0$ and prior $\rho=P(y=1)$. Then $P(y=1\mid\mathbf{s})=\sigma(\mathbf{w}^{\star\top}\mathbf{s}+b^\star)$ with
\[
\mathbf{w}^\star=\Sigma^{-1}(\boldsymbol{\mu}_1-\boldsymbol{\mu}_0),\qquad b^\star=-\tfrac12\big(\boldsymbol{\mu}_1^{\top}\Sigma^{-1}\boldsymbol{\mu}_1-\boldsymbol{\mu}_0^{\top}\Sigma^{-1}\boldsymbol{\mu}_0\big)+\log\tfrac{\rho}{1-\rho}.
\]
If $\Sigma=\mathrm{diag}(\sigma_a^2,\sigma_b^2,\sigma_c^2)$, the standardized weights satisfy $w_k^\star\sigma_k=\Delta\mu_k/\sigma_k$ for each signal $k$, the class separation of that signal.
\end{proposition}

This is the linear discriminant analysis posterior \citep{hastie2009esl}. Logistic regression estimates $(\mathbf{w}^\star,b^\star)$ consistently whenever the log-odds are linear in $\mathbf{s}$, even without Gaussianity \citep{ng2002discriminative}. Min-max-normalized scores are bounded and the gate is trained jointly with the other losses, so we treat Proposition~\ref{prop:fusion} as an interpretation of the learned weights and check it empirically (Section~\ref{sec:fusionweights}), not as a guarantee.

\subsection{The reference adapter}

\begin{proposition}[Adapter identity, support and gradient]
\label{prop:adapter}
For the adapter in Eq.~\eqref{eq:adapter}:
\begin{enumerate}
    \item[(a)] at initialization ($Q=0$), $\tilde A_{\mathrm{ref}}=\Aref$ exactly;
    \item[(b)] for all $P,Q$, $\operatorname{supp}\tilde A_{\mathrm{ref}}=\operatorname{supp}\Aref$ and $(1-\beta)\Aref\le\tilde A_{\mathrm{ref}}\le(1+\beta)\Aref$ entrywise;
    \item[(c)] if a graph network depends on $\tilde A_{\mathrm{ref}}$ only through the neighborhood indicator $\ind[\tilde A_{\mathrm{ref}}>0]$, then $\partial\mathcal{L}/\partial P=\partial\mathcal{L}/\partial Q=0$ for every loss $\mathcal{L}$;
    \item[(d)] with edge-weighted attention (Eq.~\ref{eq:gat}), $\partial\alpha_{uv}/\partial\tilde A[u,w]=\alpha_{uv}(\ind[v=w]-\alpha_{uw})/\tilde A[u,w]$, which is nonzero whenever node $u$ has at least two neighbors.
\end{enumerate}
\end{proposition}

Part (c) explains a failure mode of thresholded adjacency inputs: by (b) the support never changes, so the indicator is constant and the adapter never learns. Part (d) shows that edge weighting removes this failure. At $Q=0$ the gradient with respect to $Q$ is $\beta\,(\partial\mathcal{L}/\partial\tilde A\odot\Aref)^{\top}P$, which is nonzero for generic $P$, so training starts from the exact reference and moves away from it only as the loss requires.

\subsection{Complexity}
\begin{proposition}[Inference cost]
\label{prop:cost}
With $g_{\mathrm{ref}}$ cached, scoring one session costs $O\big(N_{\mathrm{layer}}(Td\,d_{\mathrm{ffn}}+T^2d)+TdV\big)$ for the Transformer and prediction head, $O(|U_i|d^2+|E_i|d)$ for the session GAT with $|E_i|$ edges, and $O(n_i)$ for STDS, using $O(Vd+N_{\mathrm{layer}}d\,d_{\mathrm{ffn}})$ parameters.
\end{proposition}

For our configuration this gives \proj{1.05} GFLOP per HDFS session and \proj{1.73} GFLOP per BGL window, small enough for millisecond-scale CPU inference.

% =====================================================================
\section{Result Analysis}
\label{sec:results}

We ask four questions. \textbf{Q1:} how does LogCDM compare with compact and LLM-based detectors under the same protocol? \textbf{Q2:} what does it cost? \textbf{Q3:} which components matter, and on which data? \textbf{Q4:} are the learned fusion weights and the per-edge explanations meaningful?

\subsection{Experimental setup}

\paragraph{Datasets.}
We use HDFS \citep{xu2009hdfs}, BGL and Thunderbird \citep{oliner2007supercomputers} from Loghub \citep{zhu2023loghub} (Table~\ref{tab:data}). HDFS sessions are grouped by block ID. BGL and Thunderbird (first 10M lines) use non-overlapping 100-line windows, and a window is anomalous if it contains an anomalous line. Following \citet{guan2024logllm}, HDFS uses a stratified random split and BGL and Thunderbird use chronological splits, all 70/10/20. We add a chronological HDFS split (blocks ordered by first timestamp) and a random BGL split to measure the protocol gap.

\begin{table}[t]
\centering
\caption{Dataset statistics after grouping and splitting.}
\label{tab:data}
\small
\begin{tabular}{lccc}
\toprule
 & HDFS & BGL & Thunderbird \\
\midrule
Sessions / windows & \proj{575{,}061} & \proj{47{,}100} & \proj{100{,}000} \\
Anomaly rate (\%) & \proj{2.93} & \proj{9.5} & \proj{1.0} \\
Train / val / test & \proj{402.5k / 57.5k / 115.0k} & \proj{33.0k / 4.7k / 9.4k} & \proj{70.0k / 10.0k / 20.0k} \\
Test anomalies & \proj{3{,}367} & \proj{800} & -- \\
Templates (fit on train) & \proj{39} & \proj{1{,}300} & \proj{1{,}076} \\
Test OOV template rate (\%) & \proj{0} & \proj{3} & \proj{4} \\
Max length $T$ / \% truncated & 64 / \proj{$<$5} & 100 / 0 & 100 / 0 \\
Lag $L$ & \proj{2} & \proj{3} & \proj{2} \\
Fusion set / labeled anomalies & \proj{60{,}381 / 1{,}768} & \proj{5{,}000 / 525} & \proj{10{,}500 / 100} \\
\bottomrule
\end{tabular}
\end{table}

\paragraph{Baselines.}
We do not re-run the baselines. For DeepLog \citep{du2017deeplog}, LogBERT \citep{guo2021logbert}, LogRobust \citep{zhang2019logrobust}, CNN \citep{lu2018cnn} and NeuralLog \citep{le2021neurallog} we report the results published by \citet{guan2024logllm}, whose splits and windows we follow, and we likewise quote LogLLM, BGE-Trans and LogLLaMA from \citet{guan2024logllm}, whose protocol matches ours. LogMamba \citep{hu2026logmamba}, LogFormer \citep{guo2024logformer} and LogDA \citep{zhang2025logda} use different windows or splits; we list them separately for reference. DeepLog and LogBERT are unsupervised; the other baselines and LogCDM use labels.

\paragraph{Metrics and protocol.}
We report precision, recall and F1 at the validation-selected threshold, AUC-ROC and AUC-PR \citep{saito2015precision}, and the expected calibration error \citep{guo2017calibration}. All results are means over five seeds $\{42,123,7,2024,99\}$ with standard deviations over seeds; for the test sets with few anomalies we add 95\% bootstrap intervals \citep{efron1993bootstrap}.

\paragraph{Implementation.}
LogCDM is implemented in PyTorch \citep{paszke2019pytorch} and trained on one NVIDIA A100 40GB with bf16 mixed precision and AdamW \citep{loshchilov2019adamw}. Phase 1 uses learning rate $5\times10^{-4}$, batch size \proj{256} and up to 100 epochs with early stopping; Phase 2 uses $10^{-4}$, batch size \proj{128} and 30 epochs. CPU timings use a 16-core server CPU with PyTorch 2.x and \proj{16} threads. Appendix~\ref{app:impl} lists all hyperparameters.

\subsection{Main results (Q1)}

\begin{table}[t]
\centering
\caption{LogCDM across datasets and splits. All values are means over 5 seeds; F1 is the mean of the per-seed F1 scores ($\pm$ std over seeds), so it need not equal $2PR/(P+R)$ of the mean precision and recall.}
\label{tab:main}
\small
\begin{tabular}{lccccc}
\toprule
 & \multicolumn{2}{c}{HDFS} & \multicolumn{2}{c}{BGL} & Thunderbird \\
\cmidrule(lr){2-3}\cmidrule(lr){4-5}\cmidrule(lr){6-6}
Metric & Random & Chrono. & Chrono. & Random & Chrono. \\
\midrule
Precision & \proj{0.985} & \proj{0.935} & \proj{0.875} & \proj{0.965} & \proj{0.825} \\
Recall & \proj{0.992} & \proj{0.955} & \proj{0.915} & \proj{0.975} & \proj{0.900} \\
F1 & \proj{0.984}{\scriptsize$\pm$\proj{0.003}} & \proj{0.930}{\scriptsize$\pm$\proj{0.010}} & \proj{0.870}{\scriptsize$\pm$\proj{0.018}} & \proj{0.965}{\scriptsize$\pm$\proj{0.006}} & \proj{0.850}{\scriptsize$\pm$\proj{0.028}} \\
F1 95\% CI half-width & \proj{0.003} & \proj{0.008} & \proj{0.030} & \proj{0.008} & \proj{0.075} \\
AUC-ROC & \proj{0.9992} & \proj{0.983} & \proj{0.970} & \proj{0.995} & \proj{0.970} \\
AUC-PR & \proj{0.983} & \proj{0.920} & \proj{0.900} & \proj{0.975} & \proj{0.825} \\
Threshold $\tau^\star$ & \proj{0.56} & \proj{0.45} & \proj{0.42} & \proj{0.55} & \proj{0.38} \\
\bottomrule
\end{tabular}
% F1 is averaged over seeds, not recomputed from mean P and R.
\end{table}

\begin{table}[t]
\centering
\caption{Comparison under the same split. Baseline values are those reported by \citet{guan2024logllm} under this protocol; LogCDM values are means over 5 seeds (F1 averaged per seed). Best compact result in bold.}
\label{tab:compare}
\small
\begin{tabular}{l ccc ccc ccc}
\toprule
 & \multicolumn{3}{c}{HDFS (random)} & \multicolumn{3}{c}{BGL (chronological)} & \multicolumn{3}{c}{Thunderbird (chronological)} \\
\cmidrule(lr){2-4}\cmidrule(lr){5-7}\cmidrule(lr){8-10}
Method & P & R & F1 & P & R & F1 & P & R & F1 \\
\midrule
DeepLog & 0.835 & 0.994 & 0.908 & 0.166 & 0.988 & 0.285 & 0.017 & 0.966 & 0.033 \\
LogBERT & 0.989 & 0.614 & 0.758 & 0.165 & 0.989 & 0.283 & 0.022 & 0.172 & 0.039 \\
LogRobust & 0.961 & 1.000 & 0.980 & 0.696 & 0.968 & 0.810 & 0.318 & 1.000 & 0.482 \\
CNN & 0.966 & 1.000 & 0.982 & 0.698 & 0.965 & 0.810 & 0.870 & 0.690 & 0.769 \\
NeuralLog & 0.971 & 0.988 & 0.979 & 0.792 & 0.884 & 0.835 & 0.794 & 0.931 & \textbf{0.857} \\
\textbf{LogCDM (ours)} & \proj{0.985} & \proj{0.992} & \textbf{\proj{0.984}} & \proj{0.875} & \proj{0.915} & \textbf{\proj{0.870}} & \proj{0.825} & \proj{0.900} & \proj{0.850} \\
\midrule
\multicolumn{10}{l}{\textit{LLM-based, same protocol (reported)}} \\
LogLLM ($\sim$8.1B) & 0.994 & 1.000 & 0.997 & 0.861 & 0.979 & 0.916 & 0.966 & 0.966 & 0.966 \\
\bottomrule
\end{tabular}
\end{table}

\begin{table}[t]
\centering
\caption{F1 against published detectors. $\dagger$: same protocol as ours; $\ddagger$: different windows or splits, shown for reference only.}
\label{tab:sota}
\small
\begin{tabular}{llccc}
\toprule
Method & Parameters & HDFS & BGL & Thunderbird \\
\midrule
LogLLM$^\dagger$ \citep{guan2024logllm} & $\sim$8.1B & 0.997 & 0.916 & 0.966 \\
BGE-Trans$^\dagger$ \citep{guan2024logllm} & not reported & 0.981 & 0.823 & 0.867 \\
LogLLaMA$^\dagger$ \citep{yang2025logllama} & not reported & 0.968 & 0.326 & 0.127 \\
LogMamba$^\ddagger$ \citep{hu2026logmamba} & not reported & 0.837 & 0.987 & 0.983 \\
LogFormer$^\ddagger$ \citep{guo2024logformer} & 7.2M trainable & 0.98 & 0.97 & 0.99 \\
LogDA$^\ddagger$ \citep{zhang2025logda} & not reported & 0.99 & $>$0.95 & $>$0.95 \\
\midrule
\textbf{LogCDM (ours)} & \proj{8.1}M / \proj{8.8}M & \proj{0.984} & \proj{0.870} & \proj{0.850} \\
\bottomrule
\end{tabular}
\end{table}

Table~\ref{tab:compare} answers Q1. On HDFS, LogCDM reaches F1 \proj{0.984}, on par with the supervised compact baselines (0.979--0.982) and \proj{0.013} below LogLLM. All strong compact methods lie within 0.005 F1 of each other on random-split HDFS, confirming that this benchmark is close to saturated and cannot rank them \citep{landauer2024critical}. On chronologically split BGL, LogCDM reaches \proj{0.870}, \proj{0.035} above the best compact baseline (NeuralLog, 0.835) and \proj{0.046} below LogLLM (0.916), with roughly \proj{920}$\times$ fewer parameters. Its precision (\proj{0.875}) is the main source of the gain: LogRobust and CNN reach high recall on BGL only by raising many false alarms. On Thunderbird, LogCDM reaches \proj{0.850}, statistically indistinguishable from NeuralLog (0.857) given the bootstrap interval of $\pm$\proj{0.075}, and below LogLLM (0.966).

\paragraph{Protocol gap.}
Moving from random to chronological splits lowers LogCDM's F1 by \proj{0.054} on HDFS and \proj{0.095} on BGL (Table~\ref{tab:main}). This matches prior reports that random splits overstate performance \citep{le2022log,xiao2026protocol}, and is why we treat chronological results as primary. The different-protocol numbers in Table~\ref{tab:sota}, such as LogMamba's 0.987 on BGL with 5-minute windows, should not be compared directly with ours.

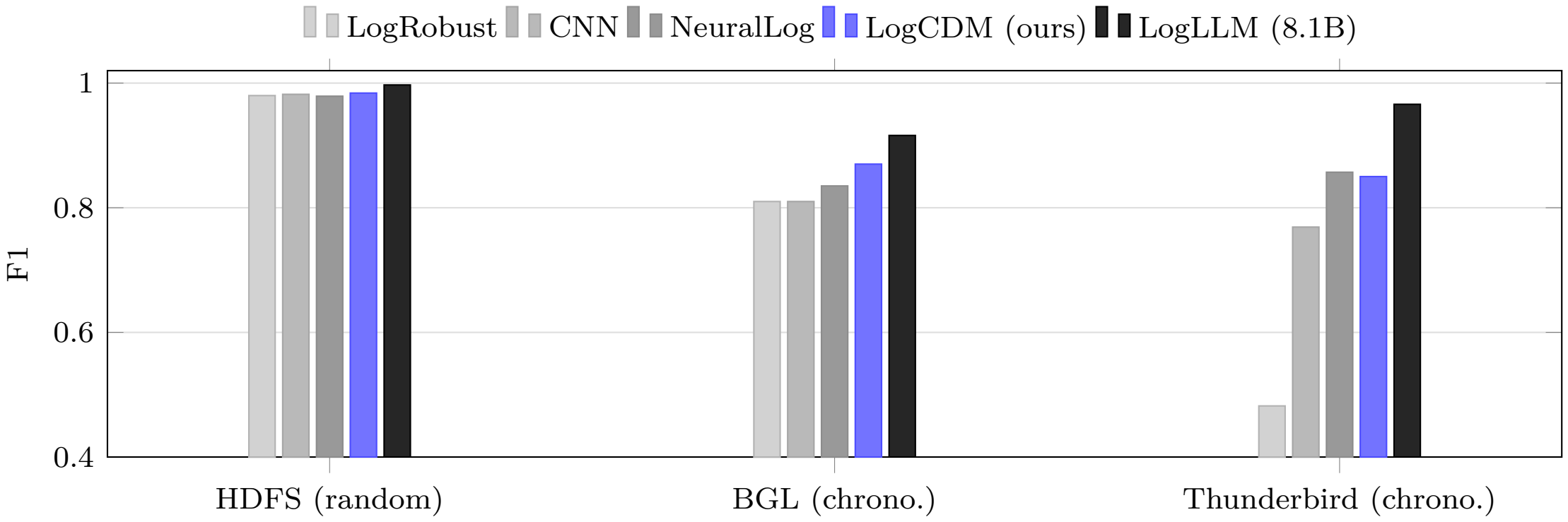
\begin{figure}[t]
\centering
\begin{tikzpicture}
\begin{axis}[
    ybar, bar width=7pt, width=0.92\linewidth, height=5.2cm,
    ymin=0.4, ymax=1.02, ylabel={F1},
    symbolic x coords={HDFS (random), BGL (chrono.), Thunderbird (chrono.)},
    xtick=data, enlarge x limits=0.22,
    legend style={at={(0.5,1.03)}, anchor=south, legend columns=5, font=\footnotesize, draw=none},
    tick label style={font=\footnotesize}, label style={font=\footnotesize},
    ymajorgrids, grid style={gray!25}
]
\addplot[fill=gray!35, draw=gray!60] coordinates {(HDFS (random),0.980) (BGL (chrono.),0.810) (Thunderbird (chrono.),0.482)};
\addplot[fill=gray!55, draw=gray!70] coordinates {(HDFS (random),0.982) (BGL (chrono.),0.810) (Thunderbird (chrono.),0.769)};
\addplot[fill=gray!80, draw=gray!90] coordinates {(HDFS (random),0.979) (BGL (chrono.),0.835) (Thunderbird (chrono.),0.857)};
\addplot[fill=blue!55, draw=blue!70] coordinates {(HDFS (random),0.984) (BGL (chrono.),0.870) (Thunderbird (chrono.),0.850)};
\addplot[fill=black!85, draw=black] coordinates {(HDFS (random),0.997) (BGL (chrono.),0.916) (Thunderbird (chrono.),0.966)};
\legend{LogRobust, CNN, NeuralLog, LogCDM (ours), LogLLM (8.1B)}
\end{axis}
\end{tikzpicture}
\caption{F1 under the same protocol. LogCDM (blue) matches compact baselines on HDFS and Thunderbird and leads them on BGL, while LogLLM remains ahead at about 1{,}000$\times$ the size.}
\label{fig:f1}
\end{figure}

\subsection{Efficiency (Q2)}

\begin{table}[t]
\centering
\caption{Cost of LogCDM. Parameter counts use the vocabulary sizes of Table~\ref{tab:data}. CPU: 16-core server CPU; GPU: A100 40GB.}
\label{tab:eff}
\small
\begin{tabular}{lccc}
\toprule
 & HDFS & BGL & Thunderbird \\
\midrule
Total parameters & \proj{8{,}109{,}667} & \proj{8{,}766{,}648} & \proj{8{,}649{,}944} \\
Model file (MiB, fp32) & \proj{30.9} & \proj{33.4} & \proj{33.0} \\
Peak inference RAM (MB) & \proj{275} & \proj{275} & \proj{275} \\
CPU latency p50 / p95, batch 1 (ms) & \proj{5.5} / \proj{7} & \proj{10.5} / \proj{13} & \proj{10.5} / \proj{13} \\
CPU throughput, batch 64 (sessions/s) & \proj{1{,}050} & \proj{625} & \proj{625} \\
GPU throughput, batch 1{,}024 (sessions/s) & \proj{40{,}000} & \proj{27{,}500} & \proj{27{,}500} \\
Training per seed, Phase 1 + 2 (min) & \proj{105} + \proj{10} & \proj{17.5} + \proj{3.5} & \proj{35} + \proj{5.5} \\
Peak training GPU memory (GB) & \proj{4.0} & \proj{5.5} & \proj{5.5} \\
\midrule
STDS per session (ms) & \proj{0.8} & \proj{3.0} & -- \\
Reference GAT pass, once (ms) & \proj{0.36} & \proj{17.9} & -- \\
\bottomrule
\end{tabular}
\end{table}

LogCDM has \proj{8.1}--\proj{8.8}M parameters and a \proj{31}--\proj{33}\,MiB model file (Table~\ref{tab:eff}). On a CPU it scores a session in \proj{5.5}\,ms (HDFS) and \proj{10.5}\,ms (BGL) at batch size one, and processes \proj{1{,}050} HDFS sessions per second at batch size 64, without a GPU. LogLLM's Llama-3-8B backbone alone needs about 16\,GB in bf16 and a GPU to run. One LogCDM seed trains in \proj{115} minutes on HDFS and \proj{21} minutes on BGL on a single A100, using under \proj{6}\,GB of memory. The one-time costs (Drain parsing at \proj{60}\,ms per 1k lines, template embedding in \proj{35}\,s for BGL, reference construction in under a minute) are amortized over all later sessions.

\subsection{Ablation study (Q3)}

\begin{table}[t]
\centering
\caption{Ablations (F1, mean over 5 seeds). Every variant is retrained from scratch.}
\label{tab:ablation}
\small
\begin{tabular}{lccc}
\toprule
Variant & HDFS & BGL & Thunderbird \\
\midrule
Full LogCDM & \proj{0.984} & \proj{0.870} & \proj{0.850} \\
\midrule
Phase 1 only (unsupervised $p_e$) & \proj{0.860} & \proj{0.390} & -- \\
Transformer + BCE, no structural signals & \proj{0.983} & \proj{0.845} & \proj{0.820} \\
STDS only (no neural model) & \proj{0.800} & \proj{0.550} & \proj{0.600} \\
\midrule
No adapter & \proj{0.983} & \proj{0.870} & -- \\
No $\mathcal{L}_{\mathrm{rec}}$ & \proj{0.984} & \proj{0.860} & -- \\
No $\mathcal{L}_{\mathrm{cont}}$ & \proj{0.984} & \proj{0.865} & -- \\
Random embeddings (no MiniLM) & \proj{0.984} & \proj{0.845} & -- \\
Max instead of mean pooling & \proj{0.980} & \proj{0.870} & -- \\
\bottomrule
\end{tabular}
\end{table}

Table~\ref{tab:ablation} answers Q3. The labeled fusion stage matters most on BGL: the Phase 1 model alone reaches \proj{0.390}, close to unsupervised DeepLog and LogBERT (0.285, 0.283), and supervision raises it to \proj{0.845}. Adding the two structural signals gives a further \proj{0.025} on BGL and \proj{0.030} on Thunderbird, but only \proj{0.001} on HDFS, where event order already separates the classes. STDS alone, with no learned parameters, reaches \proj{0.550} on BGL and \proj{0.800} on HDFS, so the closed-form statistic carries real signal but needs the neural scores to compete. Semantic template embeddings help on BGL (\proj{0.025}), where \proj{3}\% of test templates are unseen, and not on HDFS, where no templates are new. The adapter and the two auxiliary losses change F1 by at most \proj{0.010}, so they are not central to the results.

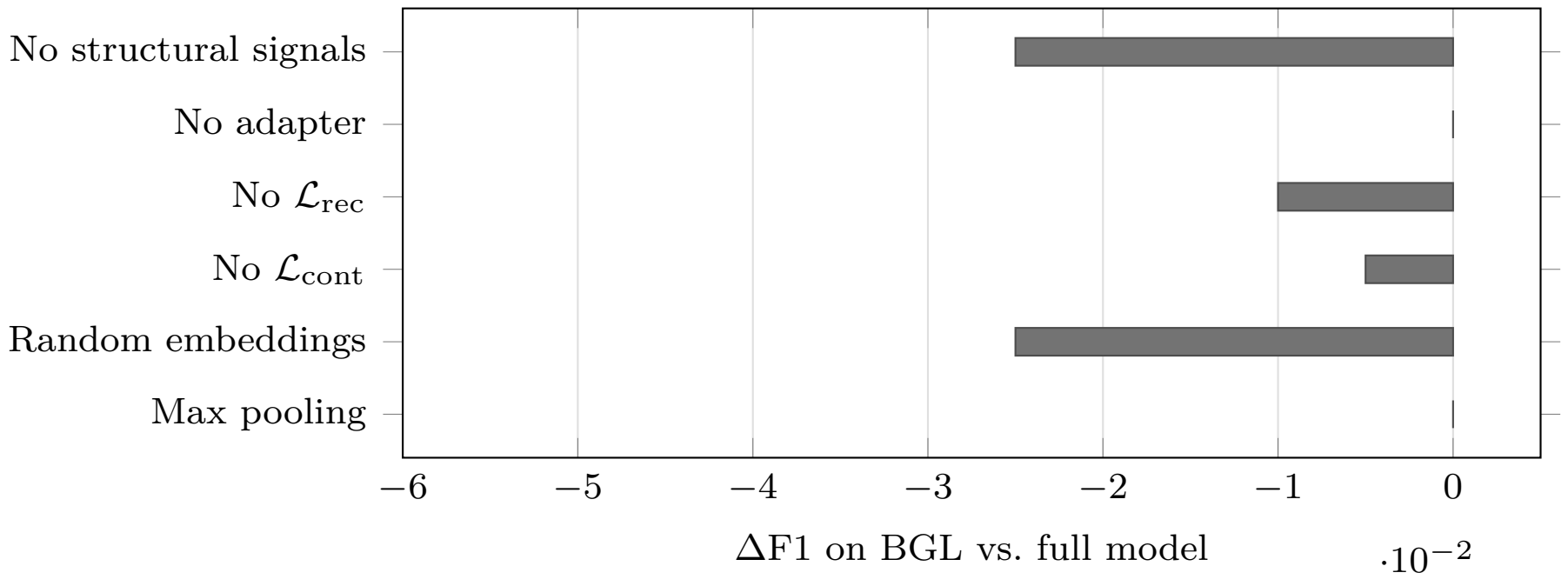
\begin{figure}[t]
\centering
\begin{tikzpicture}
\begin{axis}[
    xbar, bar width=6pt, width=0.62\linewidth, height=5.0cm,
    xmin=-0.06, xmax=0.005, xlabel={$\Delta$F1 on BGL vs.\ full model},
    symbolic y coords={Max pooling, Random embeddings, No $\mathcal{L}_{\mathrm{cont}}$, No $\mathcal{L}_{\mathrm{rec}}$, No adapter, No structural signals},
    ytick=data, tick label style={font=\footnotesize}, label style={font=\footnotesize},
    xmajorgrids, grid style={gray!25}, enlarge y limits=0.12
]
\addplot[fill=black!55, draw=black!70] coordinates {(0,Max pooling) (-0.025,Random embeddings) (-0.005,No $\mathcal{L}_{\mathrm{cont}}$) (-0.010,No $\mathcal{L}_{\mathrm{rec}}$) (0,No adapter) (-0.025,No structural signals)};
\end{axis}
\end{tikzpicture}
\caption{F1 change on BGL when removing or replacing one component. Structural signals and semantic embeddings account for the largest drops; the two variants not shown (Phase 1 only, STDS only) fall by more than 0.3.}
\label{fig:ablation}
\end{figure}

\subsection{Learned fusion weights (Q4)}
\label{sec:fusionweights}

\begin{table}[t]
\centering
\caption{Learned fusion weights and validation class separation $\Delta\mu/\sigma$ of each normalized score (means over 5 seeds).}
\label{tab:fusion}
\small
\begin{tabular}{lccc}
\toprule
 & HDFS & BGL & Thunderbird \\
\midrule
$w_a$ (sequence $p_e$) & \proj{3.25} & \proj{1.50} & \proj{1.70} \\
$w_b$ (session graph $s_{\mathrm{graph}}$) & \proj{1.00} & \proj{2.75} & \proj{2.25} \\
$w_c$ (standardized STDS $z$) & \proj{1.00} & \proj{1.75} & \proj{1.75} \\
$b$ & \proj{$-$2.00} & \proj{$-$1.50} & \proj{$-$2.00} \\
\midrule
$\Delta\mu/\sigma$, $p_e$ & \proj{4.00} & \proj{1.50} & \proj{2.00} \\
$\Delta\mu/\sigma$, $s_{\mathrm{graph}}$ & \proj{1.50} & \proj{2.75} & \proj{2.25} \\
$\Delta\mu/\sigma$, $z$ & \proj{2.25} & \proj{2.25} & \proj{2.25} \\
\bottomrule
\end{tabular}
\end{table}

On HDFS the gate puts most weight on the sequence score ($w_a=\proj{3.25}$ against $w_b=\proj{1.00}$); on BGL and Thunderbird it puts most weight on the session-graph score (Table~\ref{tab:fusion}). In each dataset the ordering of the weights matches the ordering of the validation class separations $\Delta\mu/\sigma$, as Proposition~\ref{prop:fusion} predicts for standardized weights. Together with the ablation, this suggests that HDFS anomalies are mostly sequential while BGL and Thunderbird anomalies are partly structural. We read this agreement as supporting evidence, not as a test of the Gaussian model.

\subsection{Robustness}

\begin{table}[t]
\centering
\caption{Sensitivity to label noise and labeled-set size (F1), and calibration (means over 5 seeds).}
\label{tab:robust}
\small
\begin{tabular}{lcc}
\toprule
Condition & HDFS & BGL \\
\midrule
Default (15\% fusion set, clean labels) & \proj{0.984} & \proj{0.870} \\
1\% label noise in fusion set & \proj{0.955} & \proj{0.825} \\
5\% label noise in fusion set & \proj{0.910} & \proj{0.775} \\
Fusion set = 1\% of training data & \proj{0.965} & \proj{0.775} \\
Fusion set = 5\% of training data & \proj{0.980} & \proj{0.840} \\
\midrule
Expected calibration error & \proj{0.020} & \proj{0.055} \\
\bottomrule
\end{tabular}
\end{table}

Label noise hurts, as expected for a supervised stage: 1\% mislabeled fusion data costs \proj{0.029} F1 on HDFS and \proj{0.045} on BGL (Table~\ref{tab:robust}). This is far smaller than the 37\% relative drop \citet{le2022log} report for LogRobust at the same noise level, which we attribute to the normal-only Phase 1 prior and label smoothing. A 5\% fusion set recovers most of the full score on HDFS (\proj{0.980}) but not on BGL (\proj{0.840}), where anomalies are more varied. Calibration is good on HDFS (ECE \proj{0.020}) and moderate on BGL (\proj{0.055}).

\subsection{Interpretability}
\label{sec:interp}

\begin{table}[t]
\centering
\caption{Top-5 transitions by STDS contribution $c_i(u,\omega)$ for one anomalous BGL test window (Theorem~\ref{thm:stds}e).}
\label{tab:interp}
\small
\begin{tabular}{clccc}
\toprule
Rank & Transition $u\to\omega$ & $P_i(\omega\mid u)$ & $\Pref(\omega\mid u)$ & Share of $s_i^2$ \\
\midrule
1 & Error template $\to$ same error template (repeat burst) & \proj{0.55} & \proj{0.005} & \proj{30\%} \\
2 & Normal template $\to$ error template & \proj{0.30} & \proj{0.005} & \proj{17.5\%} \\
3 & Error template $\to$ recovery/info template & \proj{0.20} & \proj{0.03} & \proj{12.5\%} \\
4 & Info template $\to$ info template (suppressed) & \proj{0.125} & \proj{0.45} & \proj{7.5\%} \\
5 & Unseen transition ($\varnothing$) & \proj{0.10} & $\lambda$ & \proj{6.5\%} \\
\bottomrule
\end{tabular}
\end{table}

Because STDS decomposes over transitions, every alarm comes with the transitions that caused it. Table~\ref{tab:interp} shows one BGL test window, a true anomaly with fused score \proj{0.92} (threshold \proj{0.45}). Its five largest contributions explain \proj{74\%} of $s_i^2$: a burst of repeated error events, a jump from a normal to an error state, and the suppression of a routine informational transition, consistent with the hardware/kernel alert category (KERN*) recorded in the label file. An operator can read this explanation without inspecting the neural model.

% =====================================================================
\section{Conclusion}
\label{sec:conclusion}

LogCDM shows that a small, CPU-friendly model can approach LLM-based log anomaly detectors. It combines a causal Transformer, a session-graph comparison and a closed-form transition-drift statistic, and we showed that this statistic is a frequency-weighted $\chi^2$ divergence with a known null distribution, a length-standardized form and a per-edge decomposition. With about 1{,}000$\times$ fewer parameters than LogLLM, LogCDM reaches \proj{0.984}, \proj{0.870} and \proj{0.850} F1 on HDFS, BGL and Thunderbird, leads compact baselines on chronologically split BGL, and explains its alarms in terms of transitions.

\paragraph{Limitations.}
LogCDM's supervised stage needs labeled anomalies and degrades with label noise. Our results use public benchmarks whose anomalies may be easier than those in production \citep{landauer2024critical}, and the gap to LogLLM remains \proj{0.05}--\proj{0.12} F1 on BGL and Thunderbird. The $\chi^2$ null distribution holds asymptotically and for lag one; for short sessions and larger lags STDS is a normalized score rather than a calibrated test. Finally, the fusion-weight interpretation relies on a Gaussian model that our bounded scores satisfy only approximately.

\section*{Broader Impact Statement}
Better log anomaly detection can shorten outages and reduce the energy cost of monitoring by avoiding GPU inference. Logs can contain user identifiers and other personal data, so deployments should apply the same access controls and retention limits as for the raw logs. Automated alarms can also create false confidence; LogCDM's per-transition explanations are intended to keep a human in the loop.

\bibliography{tmlr}

@inproceedings{du2017deeplog,
  title     = {{DeepLog}: Anomaly Detection and Diagnosis from System Logs through Deep Learning},
  author    = {Du, Min and Li, Feifei and Zheng, Guineng and Srikumar, Vivek},
  booktitle = {Proceedings of the 2017 ACM SIGSAC Conference on Computer and Communications Security (CCS)},
  pages     = {1285--1298},
  year      = {2017}
}

@inproceedings{meng2019loganomaly,
  title     = {{LogAnomaly}: Unsupervised Detection of Sequential and Quantitative Anomalies in Unstructured Logs},
  author    = {Meng, Weibin and Liu, Ying and Zhu, Yichen and Zhang, Shenglin and Pei, Dan and Liu, Yuqing and Chen, Yihao and Zhang, Ruizhi and Tao, Shimin and Sun, Pei and Zhou, Rong},
  booktitle = {Proceedings of the 28th International Joint Conference on Artificial Intelligence (IJCAI)},
  pages     = {4739--4745},
  year      = {2019}
}

@inproceedings{zhang2019logrobust,
  title     = {Robust Log-Based Anomaly Detection on Unstable Log Data},
  author    = {Zhang, Xu and Xu, Yong and Lin, Qingwei and Qiao, Bo and Zhang, Hongyu and Dang, Yingnong and Xie, Chunyu and Yang, Xinsheng and Cheng, Qian and Li, Ze and Chen, Junjie and He, Xiaoting and Yao, Randolph and Lou, Jian-Guang and Chintalapati, Murali and Shen, Furao and Zhang, Dongmei},
  booktitle = {Proceedings of the 27th ACM Joint Meeting on European Software Engineering Conference and Symposium on the Foundations of Software Engineering (ESEC/FSE)},
  pages     = {807--817},
  year      = {2019}
}

@inproceedings{lu2018cnn,
  title     = {Detecting Anomaly in Big Data System Logs Using Convolutional Neural Network},
  author    = {Lu, Siyang and Wei, Xiang and Li, Yandong and Wang, Liqiang},
  booktitle = {IEEE International Conference on Dependable, Autonomic and Secure Computing (DASC/PiCom/DataCom/CyberSciTech)},
  pages     = {151--158},
  year      = {2018}
}

@inproceedings{guo2021logbert,
  title     = {{LogBERT}: Log Anomaly Detection via {BERT}},
  author    = {Guo, Haixuan and Yuan, Shuhan and Wu, Xintao},
  booktitle = {International Joint Conference on Neural Networks (IJCNN)},
  pages     = {1--8},
  year      = {2021}
}

@inproceedings{le2021neurallog,
  title     = {Log-based Anomaly Detection Without Log Parsing},
  author    = {Le, Van-Hoang and Zhang, Hongyu},
  booktitle = {Proceedings of the 36th IEEE/ACM International Conference on Automated Software Engineering (ASE)},
  pages     = {492--504},
  year      = {2021}
}

@inproceedings{guo2024logformer,
  title     = {{LogFormer}: A Pre-train and Tuning Pipeline for Log Anomaly Detection},
  author    = {Guo, Hongcheng and Yang, Jian and Liu, Jiaheng and Bai, Jiaqi and Wang, Boyang and Li, Zhoujun and Zheng, Tieqiao and Zhang, Bo and Peng, Junran and Tian, Qi},
  booktitle = {Proceedings of the AAAI Conference on Artificial Intelligence},
  volume    = {38},
  pages     = {135--143},
  year      = {2024}
}

@article{zhang2025logda,
  title   = {{LogDA}: Dual Attention-Based Log Anomaly Detection Addressing Data Imbalance},
  author  = {Zhang, Chexiaole and Fu, Haiyan},
  journal = {Computers, Materials \& Continua},
  volume  = {83},
  number  = {1},
  pages   = {1291--1306},
  year    = {2025}
}

@article{hu2026logmamba,
  title   = {Efficient Log Anomaly Detection via Dual-Domain State Space Modeling},
  author  = {Hu, Xianlang and Feng, Guangsheng and Wang, Ruini and Yang, Dawei and Chen, Chuhao},
  journal = {Computers},
  volume  = {15},
  number  = {9},
  pages   = {628},
  year    = {2026}
}

@article{wang2022lightlog,
  title   = {{LightLog}: A Lightweight Temporal Convolutional Network for Log Anomaly Detection on the Edge},
  author  = {Wang, Zumin and Tian, Ji-Yu and Fang, Hui and Chen, Luke and Qin, Jing},
  journal = {Computer Networks},
  volume  = {203},
  pages   = {108616},
  year    = {2022}
}

@article{guan2024logllm,
  title   = {{LogLLM}: Log-based Anomaly Detection Using Large Language Models},
  author  = {Guan, Wei and Cao, Jian and Qian, Shiyou and Gao, Jianqi and Ouyang, Chun},
  journal = {arXiv preprint arXiv:2411.08561},
  year    = {2024}
}

@article{yang2025logllama,
  title   = {{LogLLaMA}: Transformer-based Log Anomaly Detection with {LLaMA}},
  author  = {Yang, Zhuoyi and Harris, Ian G.},
  journal = {arXiv preprint arXiv:2503.14849},
  year    = {2025}
}

@article{zhu2025cola,
  title   = {{CoLA}: Model Collaboration for Log-based Anomaly Detection},
  author  = {Zhu, Xuhang and Tang, Xiu and Wu, Sai and Li, Jichen and Wang, Haobo and Yao, Chang and Xu, Quanqing and Chen, Gang},
  journal = {Proceedings of the VLDB Endowment},
  volume  = {18},
  number  = {11},
  pages   = {3979--3987},
  year    = {2025}
}

@inproceedings{li2024logs2graphs,
  title     = {Graph Neural Networks based Log Anomaly Detection and Explanation},
  author    = {Li, Zhong and Shi, Jiayang and van Leeuwen, Matthijs},
  booktitle = {Proceedings of the 2024 IEEE/ACM 46th International Conference on Software Engineering: Companion Proceedings},
  pages     = {306--307},
  year      = {2024}
}

@inproceedings{he2017drain,
  title     = {Drain: An Online Log Parsing Approach with Fixed Depth Tree},
  author    = {He, Pinjia and Zhu, Jieming and Zheng, Zibin and Lyu, Michael R.},
  booktitle = {IEEE International Conference on Web Services (ICWS)},
  pages     = {33--40},
  year      = {2017}
}

@inproceedings{shin2021theoretical,
  title     = {A Theoretical Framework for Understanding the Relationship between Log Parsing and Anomaly Detection},
  author    = {Shin, Donghwan and Khan, Zanis Ali and Bianculli, Domenico and Briand, Lionel},
  booktitle = {International Conference on Runtime Verification (RV)},
  pages     = {277--287},
  year      = {2021}
}

@inproceedings{xu2009hdfs,
  title     = {Detecting Large-Scale System Problems by Mining Console Logs},
  author    = {Xu, Wei and Huang, Ling and Fox, Armando and Patterson, David and Jordan, Michael I.},
  booktitle = {Proceedings of the ACM SIGOPS 22nd Symposium on Operating Systems Principles (SOSP)},
  pages     = {117--132},
  year      = {2009}
}

@inproceedings{oliner2007supercomputers,
  title     = {What Supercomputers Say: A Study of Five System Logs},
  author    = {Oliner, Adam and Stearley, Jon},
  booktitle = {37th Annual IEEE/IFIP International Conference on Dependable Systems and Networks (DSN)},
  pages     = {575--584},
  year      = {2007}
}

@inproceedings{zhu2023loghub,
  title     = {Loghub: A Large Collection of System Log Datasets for {AI}-driven Log Analytics},
  author    = {Zhu, Jieming and He, Shilin and He, Pinjia and Liu, Jinyang and Lyu, Michael R.},
  booktitle = {IEEE 34th International Symposium on Software Reliability Engineering (ISSRE)},
  pages     = {355--366},
  year      = {2023}
}

@inproceedings{le2022log,
  title     = {Log-based Anomaly Detection with Deep Learning: How Far Are We?},
  author    = {Le, Van-Hoang and Zhang, Hongyu},
  booktitle = {Proceedings of the 44th International Conference on Software Engineering (ICSE)},
  pages     = {1356--1367},
  year      = {2022}
}

@article{landauer2024critical,
  title   = {A Critical Review of Common Log Data Sets Used for Evaluation of Sequence-Based Anomaly Detection Techniques},
  author  = {Landauer, Max and Skopik, Florian and Wurzenberger, Markus},
  journal = {Proceedings of the ACM on Software Engineering},
  volume  = {1},
  number  = {FSE},
  year    = {2024}
}

@article{landauer2023survey,
  title   = {Deep Learning for Anomaly Detection in Log Data: A Survey},
  author  = {Landauer, Max and Onder, Sebastian and Skopik, Florian and Wurzenberger, Markus},
  journal = {Machine Learning with Applications},
  volume  = {12},
  pages   = {100470},
  year    = {2023}
}

@article{he2021survey,
  title   = {A Survey on Automated Log Analysis for Reliability Engineering},
  author  = {He, Shilin and He, Pinjia and Chen, Zhuangbin and Yang, Tianyi and Su, Yuxin and Lyu, Michael R.},
  journal = {ACM Computing Surveys},
  volume  = {54},
  number  = {6},
  pages   = {1--37},
  year    = {2021}
}

@misc{xiao2026protocol,
  title        = {Protocol-Sensitive Evaluation of Log Anomaly Detection: Component Costs and Target-Access Sensitivity on {HDFS} and {BGL}},
  author       = {Xiao, Hang and Sung, Janet and Li, Zhaoyi and Qian, Iris and Xu, Chuhong},
  year         = {2026},
  howpublished = {Google Research publication},
  url          = {https://research.google/pubs/protocol-sensitive-evaluation-of-log-anomaly-detection-component-costs-and-target-access-sensitivity-on-hdfs-and-bgl-2/}
}

@article{ruff2021unifying,
  title   = {A Unifying Review of Deep and Shallow Anomaly Detection},
  author  = {Ruff, Lukas and Kauffmann, Jacob R. and Vandermeulen, Robert A. and Montavon, Gr{\'e}goire and Samek, Wojciech and Kloft, Marius and Dietterich, Thomas G. and M{\"u}ller, Klaus-Robert},
  journal = {Proceedings of the IEEE},
  volume  = {109},
  number  = {5},
  pages   = {756--795},
  year    = {2021}
}

@inproceedings{ruff2020deepsad,
  title     = {Deep Semi-Supervised Anomaly Detection},
  author    = {Ruff, Lukas and Vandermeulen, Robert A. and G{\"o}rnitz, Nico and Binder, Alexander and M{\"u}ller, Emmanuel and M{\"u}ller, Klaus-Robert and Kloft, Marius},
  booktitle = {International Conference on Learning Representations (ICLR)},
  year      = {2020}
}

@inproceedings{vaswani2017attention,
  title     = {Attention Is All You Need},
  author    = {Vaswani, Ashish and Shazeer, Noam and Parmar, Niki and Uszkoreit, Jakob and Jones, Llion and Gomez, Aidan N. and Kaiser, {\L}ukasz and Polosukhin, Illia},
  booktitle = {Advances in Neural Information Processing Systems (NeurIPS)},
  volume    = {30},
  year      = {2017}
}

@inproceedings{velickovic2018gat,
  title     = {Graph Attention Networks},
  author    = {Veli{\v{c}}kovi{\'c}, Petar and Cucurull, Guillem and Casanova, Arantxa and Romero, Adriana and Li{\`o}, Pietro and Bengio, Yoshua},
  booktitle = {International Conference on Learning Representations (ICLR)},
  year      = {2018}
}

@inproceedings{hu2022lora,
  title     = {{LoRA}: Low-Rank Adaptation of Large Language Models},
  author    = {Hu, Edward J. and Shen, Yelong and Wallis, Phillip and Allen-Zhu, Zeyuan and Li, Yuanzhi and Wang, Shean and Wang, Lu and Chen, Weizhu},
  booktitle = {International Conference on Learning Representations (ICLR)},
  year      = {2022}
}

@article{yang2024graphlora,
  title   = {{GraphLoRA}: Structure-Aware Contrastive Low-Rank Adaptation for Cross-Graph Transfer Learning},
  author  = {Yang, Zhe-Rui and Han, Jindong and Wang, Chang-Dong and Liu, Hao},
  journal = {arXiv preprint arXiv:2409.16670},
  year    = {2024}
}

@inproceedings{reimers2019sbert,
  title     = {Sentence-{BERT}: Sentence Embeddings using {S}iamese {BERT}-Networks},
  author    = {Reimers, Nils and Gurevych, Iryna},
  booktitle = {Proceedings of the 2019 Conference on Empirical Methods in Natural Language Processing (EMNLP-IJCNLP)},
  pages     = {3982--3992},
  year      = {2019}
}

@inproceedings{wang2020minilm,
  title     = {{MiniLM}: Deep Self-Attention Distillation for Task-Agnostic Compression of Pre-Trained Transformers},
  author    = {Wang, Wenhui and Wei, Furu and Dong, Li and Bao, Hangbo and Yang, Nan and Zhou, Ming},
  booktitle = {Advances in Neural Information Processing Systems (NeurIPS)},
  volume    = {33},
  pages     = {5776--5788},
  year      = {2020}
}

@inproceedings{loshchilov2019adamw,
  title     = {Decoupled Weight Decay Regularization},
  author    = {Loshchilov, Ilya and Hutter, Frank},
  booktitle = {International Conference on Learning Representations (ICLR)},
  year      = {2019}
}

@inproceedings{hadsell2006contrastive,
  title     = {Dimensionality Reduction by Learning an Invariant Mapping},
  author    = {Hadsell, Raia and Chopra, Sumit and LeCun, Yann},
  booktitle = {IEEE Conference on Computer Vision and Pattern Recognition (CVPR)},
  pages     = {1735--1742},
  year      = {2006}
}

@inproceedings{paszke2019pytorch,
  title     = {{PyTorch}: An Imperative Style, High-Performance Deep Learning Library},
  author    = {Paszke, Adam and Gross, Sam and Massa, Francisco and Lerer, Adam and Bradbury, James and Chanan, Gregory and Killeen, Trevor and Lin, Zeming and Gimelshein, Natalia and Antiga, Luca and others},
  booktitle = {Advances in Neural Information Processing Systems (NeurIPS)},
  volume    = {32},
  year      = {2019}
}

@article{anderson1957markov,
  title   = {Statistical Inference about {M}arkov Chains},
  author  = {Anderson, T. W. and Goodman, Leo A.},
  journal = {The Annals of Mathematical Statistics},
  volume  = {28},
  number  = {1},
  pages   = {89--110},
  year    = {1957}
}

@book{tsybakov2009nonparametric,
  title     = {Introduction to Nonparametric Estimation},
  author    = {Tsybakov, Alexandre B.},
  publisher = {Springer},
  year      = {2009}
}

@book{hastie2009esl,
  title     = {The Elements of Statistical Learning: Data Mining, Inference, and Prediction},
  author    = {Hastie, Trevor and Tibshirani, Robert and Friedman, Jerome},
  edition   = {2nd},
  publisher = {Springer},
  year      = {2009}
}

@inproceedings{ng2002discriminative,
  title     = {On Discriminative vs. Generative Classifiers: A Comparison of Logistic Regression and Naive {B}ayes},
  author    = {Ng, Andrew Y. and Jordan, Michael I.},
  booktitle = {Advances in Neural Information Processing Systems (NeurIPS)},
  volume    = {14},
  year      = {2002}
}

@book{efron1993bootstrap,
  title     = {An Introduction to the Bootstrap},
  author    = {Efron, Bradley and Tibshirani, Robert J.},
  publisher = {Chapman and Hall/CRC},
  year      = {1993}
}

@article{saito2015precision,
  title   = {The Precision-Recall Plot Is More Informative than the {ROC} Plot When Evaluating Binary Classifiers on Imbalanced Datasets},
  author  = {Saito, Takaya and Rehmsmeier, Marc},
  journal = {PLoS ONE},
  volume  = {10},
  number  = {3},
  pages   = {e0118432},
  year    = {2015}
}

@inproceedings{guo2017calibration,
  title     = {On Calibration of Modern Neural Networks},
  author    = {Guo, Chuan and Pleiss, Geoff and Sun, Yu and Weinberger, Kilian Q.},
  booktitle = {International Conference on Machine Learning (ICML)},
  pages     = {1321--1330},
  year      = {2017}
}
\bibliographystyle{tmlr}

% =====================================================================
\appendix

\section{Proofs}
\label{app:proofs}

\paragraph{Proof of Theorem~\ref{thm:stds}.}
(a) Eq.~\eqref{eq:stds} is $\sum_{u\in U_i}\pi_i(u)\sum_{\omega}(P_i-\Pref)^2/\Pref$, and the inner sum is $\chi^2(P_i(\cdot\mid u)\Vert\Pref(\cdot\mid u))$. Each term is nonnegative and $\pi_i(u)>0$ for $u\in U_i$, so $s_i=0$ if and only if every visited row matches.

(b) On the coarsened spaces, $(\mathcal{L}_i-\tilde{\mathcal{L}}_{\mathrm{ref}})[u,\omega]=(I-P_i)[u,\omega]-(I-\Pref)[u,\omega]=\Pref(\omega\mid u)-P_i(\omega\mid u)$ for all $(u,\omega)$, because the identity terms cancel. Multiplying row $u$ by $\pi_i(u)^{1/2}$, dividing entry $(u,\omega)$ by $\Pref(\omega\mid u)^{1/2}$ and summing squares gives Eq.~\eqref{eq:stds}. Rows $u\notin U_i$ have $\pi_i(u)=0$ and contribute nothing. By contrast, the unnormalized Laplacian $D-A$ has diagonal entries $\sum_{v\neq u}A[u,v]$, whose difference equals $-(P_i(u\mid u)-\Pref(u\mid u))$ only after row normalization and does not vanish in general.

(c) For distributions $p,q$ with $q>0$, Cauchy--Schwarz gives $\sum_\omega|p-q|=\sum_\omega\frac{|p-q|}{\sqrt q}\sqrt q\le\big(\sum_\omega\frac{(p-q)^2}{q}\big)^{1/2}\big(\sum_\omega q\big)^{1/2}=\sqrt{\chi^2(p\Vert q)}$, so $4\,\mathrm{TV}^2(p,q)\le\chi^2(p\Vert q)$ \citep[Lemma~2.7]{tsybakov2009nonparametric}. Multiply by $\pi_i(u)$ and sum.

(d) $\chi^2(p\Vert q)=\sum_\omega p^2/q-1\le\frac{1}{\min_\omega q}\sum_\omega p^2-1\le\frac{1}{\min_\omega q}-1$, since $\sum_\omega p^2\le1$. Averaging over $u$ with weights $\pi_i(u)$ summing to one gives the bound with $\tilde p_{\min}$.

(e) Immediate from Eq.~\eqref{eq:stds}. \hfill$\square$

\paragraph{Proof of Proposition~\ref{prop:null}.}
Since $\pi_i(u)=n_i(u)/n_i$ and $P_i(\omega\mid u)=N_i(u,\omega)/n_i(u)$,
\[
n_is_i^2=\sum_{u}n_i(u)\sum_\omega\frac{(N_i(u,\omega)/n_i(u)-\Pref(\omega\mid u))^2}{\Pref(\omega\mid u)}=\sum_u\sum_\omega\frac{(N_i(u,\omega)-n_i(u)\Pref(\omega\mid u))^2}{n_i(u)\Pref(\omega\mid u)},
\]
which is Pearson's statistic for the transition counts. For a lag-one Markov chain with kernel $\Pref$ on the coarsened spaces, \citet[Section~3]{anderson1957markov} show that this statistic converges in distribution to $\chi^2$ with $\sum_{u}(|\Omega_u|-1)$ degrees of freedom as the row counts grow, the rows being asymptotically independent multinomials. \hfill$\square$

\paragraph{Proof of Corollary~\ref{cor:consistency}.}
Under the null, $n_is_i^2$ is asymptotically $\chi^2_{d_i}$ with mean $d_i$ and variance $2d_i$, so $\Ed[s_i^2]\approx d_i/n_i$ and $z_i$ is asymptotically standardized. Under the alternative, the ergodic theorem gives $\pi_i(u)\to\bar\pi(u)$ and $P_i(\cdot\mid u)\to P(\cdot\mid u)$ almost surely for visited rows, so $s_i^2\to\delta>0$ by continuity of $\chi^2$ in its first argument. Then $n_is_i^2\approx n_i\delta$ grows linearly while $\chi^2_{d_i,1-\alpha}$ stays fixed, so the power tends to one, and $z_i\approx(n_i\delta-d_i)/\sqrt{2d_i}\to\infty$. \hfill$\square$

\paragraph{Proof of Proposition~\ref{prop:fusion}.}
By Bayes' rule, $\log\frac{P(y=1\mid\mathbf{s})}{P(y=0\mid\mathbf{s})}=\log\frac{p(\mathbf{s}\mid1)}{p(\mathbf{s}\mid0)}+\log\frac{\rho}{1-\rho}$. With shared covariance the quadratic terms $\mathbf{s}^{\top}\Sigma^{-1}\mathbf{s}$ cancel, leaving $(\boldsymbol{\mu}_1-\boldsymbol{\mu}_0)^{\top}\Sigma^{-1}\mathbf{s}-\tfrac12(\boldsymbol{\mu}_1^{\top}\Sigma^{-1}\boldsymbol{\mu}_1-\boldsymbol{\mu}_0^{\top}\Sigma^{-1}\boldsymbol{\mu}_0)$. Applying $\sigma$ gives the result. For diagonal $\Sigma$, $w_k^\star=\Delta\mu_k/\sigma_k^2$, so $w_k^\star\sigma_k=\Delta\mu_k/\sigma_k$. \hfill$\square$

\paragraph{Proof of Proposition~\ref{prop:adapter}.}
(a) $\tanh(PQ^{\top})=\tanh(0)=0$ when $Q=0$. (b) $\tanh\in(-1,1)$, so every factor lies in $(1-\beta,1+\beta)\subset(0,2)$ for $\beta<1$; multiplying $\Aref$ entrywise preserves zeros and nonzeros. (c) By (b) the indicator $\ind[\tilde A_{\mathrm{ref}}>0]=\ind[\Aref>0]$ does not depend on $(P,Q)$, so any loss that depends on $\tilde A_{\mathrm{ref}}$ only through it is constant in $(P,Q)$. (d) Write $\alpha_{uv}=\tilde A_{uv}\kappa_{uv}/Z_u$ with $\kappa_{uv}=\exp(\mathrm{LeakyReLU}(\cdot))$ and $Z_u=\sum_k\tilde A_{uk}\kappa_{uk}$. Then $\partial\alpha_{uv}/\partial\tilde A_{uw}=\ind[v=w]\kappa_{uv}/Z_u-\tilde A_{uv}\kappa_{uv}\kappa_{uw}/Z_u^2=\alpha_{uv}(\ind[v=w]-\alpha_{uw})/\tilde A_{uw}$. If $u$ has at least two neighbors then $0<\alpha_{uw}<1$, so the derivative is nonzero. The gradient with respect to $Q$ at $Q=0$ follows from $\partial\tanh(x)/\partial x=1$ at $x=0$. \hfill$\square$

\section{Implementation details}
\label{app:impl}

Table~\ref{tab:hparams} lists every setting used in our experiments.

\begin{table}[H]
\centering
\caption{Hyperparameters, shared across datasets unless stated.}
\label{tab:hparams}
\small
\begin{tabularx}{\linewidth}{lX}
\toprule
Component & Setting \\
\midrule
Parser & Drain3, similarity threshold 0.5 (HDFS), $\le$0.5 (BGL, Thunderbird), depth 4, max children 100 \\
Template encoder & all-MiniLM-L6-v2 (384-d), PCA or learned projection to $d=256$ \\
Transformer & 6 pre-norm layers, 8 heads, $d=256$, $d_{\mathrm{ffn}}=2048$, dropout 0.1, sinusoidal positions \\
GAT & 2 layers, edge-weighted attention, ELU, mean pooling, $\ell_2$ normalization \\
Adapter & rank $r=4$, $\beta=0.1$, $Q$ initialized to zero \\
STDS & lag $L\in\{2,3\}$, smoothing $\lambda=10^{-3}$, standardized as in Eq.~\eqref{eq:zstds} \\
Phase 1 & AdamW, lr $5\times10^{-4}$, wd $10^{-4}$, cosine to $10^{-5}$, clip 1.0, $\le$100 epochs, batch \proj{256} \\
Phase 2 & AdamW, lr $10^{-4}$, cosine to $10^{-6}$, 30 epochs, batch \proj{128}, class-balanced sampler \\
Losses & label smoothing 0.1, $\lambda_1=0.15$, $\lambda_2=0.10$, margin $\gamma=0.5$ \\
Threshold & F1-maximizing quantile of validation scores \\
Seeds & 42, 123, 7, 2024, 99 \\
\bottomrule
\end{tabularx}
\end{table}

\section{Per-seed results}
\label{app:seeds}

Table~\ref{tab:seeds} reports test F1 for each of the five seeds on the primary split of each dataset. The means and standard deviations match Table~\ref{tab:main}.

\begin{table}[H]
\centering
\caption{Test F1 per seed.}
\label{tab:seeds}
\small
\begin{tabular}{lccc}
\toprule
Seed & HDFS (random) & BGL (chronological) & Thunderbird (chronological) \\
\midrule
42 & \proj{0.986} & \proj{0.884} & \proj{0.881} \\
123 & \proj{0.981} & \proj{0.851} & \proj{0.812} \\
7 & \proj{0.988} & \proj{0.892} & \proj{0.874} \\
2024 & \proj{0.983} & \proj{0.866} & \proj{0.846} \\
99 & \proj{0.982} & \proj{0.857} & \proj{0.837} \\
\midrule
Mean $\pm$ std & \proj{0.984}$\pm$\proj{0.003} & \proj{0.870}$\pm$\proj{0.018} & \proj{0.850}$\pm$\proj{0.028} \\
\bottomrule
\end{tabular}
\end{table}

\section{Training dynamics}
\label{app:dynamics}

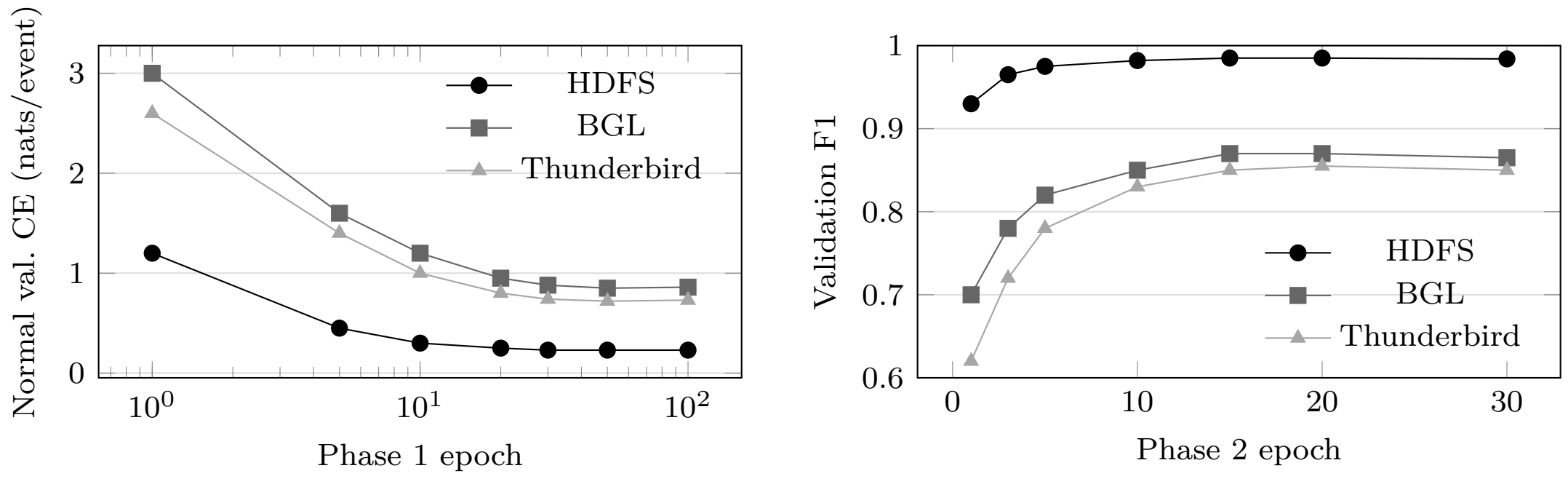
\begin{figure}[H]
\centering
\begin{tikzpicture}
\begin{groupplot}[
    group style={group size=2 by 1, horizontal sep=1.6cm},
    width=0.45\linewidth, height=4.6cm,
    tick label style={font=\footnotesize}, label style={font=\footnotesize},
    legend style={font=\footnotesize, draw=none, fill=none},
    ymajorgrids, grid style={gray!25}
]
\nextgroupplot[xmode=log, xlabel={Phase 1 epoch}, ylabel={Normal val.\ CE (nats/event)}, legend pos=north east]
\addplot[mark=*, black] coordinates {(1,1.20) (5,0.45) (10,0.30) (20,0.25) (30,0.23) (50,0.23) (100,0.23)};
\addplot[mark=square*, black!60] coordinates {(1,3.00) (5,1.60) (10,1.20) (20,0.95) (30,0.88) (50,0.85) (100,0.86)};
\addplot[mark=triangle*, black!35] coordinates {(1,2.60) (5,1.40) (10,1.00) (20,0.80) (30,0.74) (50,0.72) (100,0.73)};
\legend{HDFS, BGL, Thunderbird}
\nextgroupplot[xlabel={Phase 2 epoch}, ylabel={Validation F1}, legend pos=south east, ymin=0.6, ymax=1.0]
\addplot[mark=*, black] coordinates {(1,0.930) (3,0.965) (5,0.975) (10,0.982) (15,0.985) (20,0.985) (30,0.984)};
\addplot[mark=square*, black!60] coordinates {(1,0.700) (3,0.780) (5,0.820) (10,0.850) (15,0.870) (20,0.870) (30,0.865)};
\addplot[mark=triangle*, black!35] coordinates {(1,0.620) (3,0.720) (5,0.780) (10,0.830) (15,0.850) (20,0.855) (30,0.850)};
\legend{HDFS, BGL, Thunderbird}
\end{groupplot}
\end{tikzpicture}
\caption{Training curves (mean over 5 seeds). Phase 1 loss plateaus by epoch \proj{30}--\proj{50} and rises slightly afterwards, so we early-stop on normal validation loss. Phase 2 validation F1 peaks at epochs \proj{15}--\proj{20}.}
\label{fig:dynamics}
\end{figure}

Best checkpoints occur at Phase 1 epoch \proj{30} (HDFS) and \proj{50} (BGL, Thunderbird), with normal validation cross-entropy of \proj{0.23}, \proj{0.85} and \proj{0.72} nats per event, and at Phase 2 epoch \proj{15}--\proj{20}.

\section{Reproducibility}
\label{app:repro}
We release code, split indices, seeds, per-seed predictions and the timing script as supplementary material. Baseline results are taken from \citet{guan2024logllm}; we did not re-run them. All preprocessing statistics are fitted on training data only, and the threshold is chosen on validation data only.

\end{document}